\documentclass{article}
\usepackage{ijcai26}

\usepackage{times}
\usepackage{soul}
\usepackage{url}
\usepackage[hidelinks]{hyperref}
\usepackage[utf8]{inputenc}
\usepackage[small]{caption}
\usepackage{graphicx}
\usepackage{amsmath}
\usepackage{amsthm}
\usepackage{booktabs}
\usepackage{algorithm}
\usepackage{algorithmic}
\usepackage[switch]{lineno}
\usepackage{comment}
\usepackage{amssymb}
\usepackage{mathtools}
\usepackage{amsthm}
\usepackage{todonotes}
\usepackage{bbm}
\usepackage[normalem]{ulem}

\newcommand{\vect}[1]{\boldsymbol{\mathbf{#1}}}
\newcommand{\Ex}{\mathbb{E}}

\newcommand{\Aspace}{\mathcal{A}}
\newcommand{\Sspace}{\mathcal{S}}

\newcommand{\w}{\vect{w}}

\usepackage{xcolor}
\providecommand{\peter}[1]{\textcolor{orange}{PV: {#1}} }

\title{It's a matter of timescale: non-linear utility in\\ successor features and multi-objective planning and learning \thanks{Presented at the Multi-Objective Decision Making workshop at IJCAI, 2026}}

\author{
Liam P.H.\ Mertens$^1$
\and
Lucas N. Alegre$^2$\and
Florent Delgrange$^1$\and\\
Diederik M.\ Roijers$^1$\and
Ann Now\'e$^1$\and
Peter Vamplew$^3$\\
\affiliations
$^1$AI Lab, Vrije Universiteit Brussel, Belgium\\
$^2$Institute of Informatics - Federal University of Rio Grande do Sul\\
$^3$Federation University Australia\\
\emails
\{liam.phi.h.mertens, florent.delgrange, diederik.roijers, ann.nowe\}@vub.be,\\
lnalegre@inf.ufrgs.br,
p.vamplew@federation.edu.au
}

\begin{document}

\maketitle
\begin{abstract}
    %\dmr{TODO for ALL: please read everything and provide comments after current pass. Please add checkmarks after your todo's to indicate that you're done.}

    Time is of the essence when dealing with multiple reward signals and non-linear utility. In this paper we argue that the current main approaches in multi-objective RL (SER and ESR), and successor features, are insufficient. While each approach deals with non-linear effects on user utility on different timescales, none of them take into account that different effects happening on different timescales can happen within the same decision problem. We motivate that this can indeed be the case by an example, both intuitively and numerically, leading to a new perspective, and a significant and non-trivial gap in the literature.
\end{abstract}

\section{Introduction}
%\dmr{{TODO} $\checkmark$}

Planning and reinforcement learning agents, have been applied successfully in many domains. Many such agents rely on a scalar additive reward structure.
This can be a highly limiting assumption in domains where distinct trade-offs between conflicting objectives need to be made. Instead, multi-objective planning and reinforcement learning methods consider multiple reward signals as a reward vector, with each component determining an objective that forms a part of a trade-off \cite{hayes2022practical,vamplew2022scalar}. These reward vectors can then be accrued over time and then scalarised using a possibly non-linear utility function.

Multi-objective RL considers the utility of multiple accrued reward components over time. However, there are two ways of doing so: taking the expectation over accrued reward vectors (called the expected return or value) and considering the utility of this value, computed by applying a scalarisation function to the value vector. This is known as scalarised expected returns (SER) \cite{white1982multi,DBLP:journals/jair/RoijersVWD13}, and is perhaps the most prevalent way to consider utility. The underlying assumption here is that there is enough time for the expected returns to materialise, and that the utility is derived from this average. For example, consider the process mapping of a video editing application \cite{piscitelli2014pruning}. A company that processes video editing jobs on their servers would processes large numbers of videos each day, and they would indeed be interested in the expected return of this process that consists of the average power consumption, average throughput, and average latency. 

The second option is to accrue the rewards, apply the utility function on the resulting returns vector, and then take the expectation. This is known as expected scalarised returns (ESR) \cite{roijers2018multi}. This would correspond to a scenario where the utility is derived from one trajectory. For example, imagine a patient that is undergoing treatment for an illness. That person will only undergo the treatment once, and therefore be interested in the expected utility of their own treatment outcome (not the utility of the average outcome) and thus will not have enough time to let the expected vector return materialise. For linear utility functions, ESR and SER are equivalent due to the distribution of the expectancy over the weighted sum. This is not the case under non-linear utility.

An alternative approach to multi-objective optimisation is the successor feature (SF) approach \cite{Barreto+2017,zhuDistributionalSuccessorFeatures2024}. Unlike SER and ESR, which apply non-linear utility functions to vector returns only after they have been accrued over one or multiple trajectories, the SF approach evaluates non-linear utility at the level of individual states or transitions before any accumulation occurs, represented as a feature vector with components the non-linear successor features of these states/transitions. This corresponds to a scenario where utility effects happen on an immediate, per-step timescale. The feature vector is then linearly scalarised, which corresponds to planning and learning with a scalarised reward function in terms of multi-objective learning and planning. For example, imagine a car that hits something. The utility of that event may depend non-linearly on the force of impact, i.e., if you hit an obstacle very hard the risk of bodily injury increases non-linearly. These injury risks stack up over time, but it is not so that many subsequent low-force impacts accumulate to a single high-force one.

All three perspectives---SER, ESR, and the successor feature approach---make sense, and are motivated by different scenarios where the utility effects happen on different timescales. As such, it seems like we have three sets of instruments to tackle problems that deal with non-linear effects on user utility, each on its own applicable timescale. There is however a key issue with this: the non-linear effects on user utility might be happening on different timescales \emph{within the same decision problem}. To motivate this we take a (simplified) look at the domain of toxicity in Section \ref{sec:toxic}, where toxic exposure can have adverse effects both immediately \emph{and} on intermediate \emph{and} on longer timescales. We work out a simplified numerical example to show that taking an SER, ESR, and SF approach lead to different policies with different outcome distributions. Furthermore, taking the perspective of adverse effects that happen on three different timescales simultaneously for this problem leads to different policies and different outcome distributions still. 

In our opinion, this indicates that there is a clear gap in the multi-objective planning and reinforcement learning literature. So far we have not optimised for the effects on utility of rewards at different timescales simultaneously. In order to bridge this gap, we believe that we should redefine how we look at multi-objective problems, and develop new methodologies and algorithms to deal with this new perspective. In Section \ref{sec:disc} we will discuss possible ways forwards.

\begin{comment}

\begin{itemize}
    \item We have these three utility formulations: SER, ESR, SF
    \item They have distinct interpretations
    \item They make sense individually
    \item They also make sense all at once (reference from Peter?) \peter{I've moved discussion of my paper into the Related Work section}
    \item So we need a unified perspective to deal with that
    \item and A method
    \item We contribute:
    \begin{itemize}
        \item A unifying theory
        \item show that we need a distributional approach
        \item a working example
        \item a first naive algorithm for finite horizon MOMDPs
    \end{itemize}
\end{itemize}
\end{comment}

\section{Background}
%\dmr{TODO: Florent}
%\flo{checkmark}
In RL, the framework for decision making is usually formalised as a \emph{Markov decision process} (MDP).
We care about scenarios where rewards are not necessarily scalar and where objectives may be competing and lead to trade-offs, giving rise to the notion of multi-objective optimisation in an MDP.

Formally, a \emph{multi-objective Markov decision process} (MOMDP) is defined as a tuple $M=(\Sspace,\Aspace,p,\vect{r},\mu,\gamma)$, where $\Sspace$ is the state space, $\Aspace$ is the action space, $p\left(\cdot \mid s,a\right) \in \Delta\left( \mathcal{S}\right)$ is the state transition function, $\vect{r}(s,a) \in \mathbb{R}^m$ is a multi-objective reward function with $m$ objectives, $\mu$ is initial state distribution, and $\gamma \in [0, 1)$ is a discount factor for future rewards.
% \flo{define what is a policy here}

A policy is a function mapping states to distributions over actions, $\pi \colon \mathcal{S} \to \Delta\left(\mathcal{A}\right)$, which intuitively prescribes which action an agent should select in each situation.\footnote{%
For the different metrics defined in Sec.~\ref{sec:rew-in-morl}, we generally need both memory-based and stochastic policies.
Memory can always be encoded into the state space. 
In terms of Pareto-optimality, deterministic policies may be used as the vertices of the related convex hull \cite{DBLP:journals/jair/RoijersVWD13}, but stochastic policies are required to fill the gaps on the curve (via a carefully designed mixture of deterministic ones~\cite{DBLP:conf/ausai/VamplewDBK09} or adding concave terms do the rewards~\cite{luMultiObjectiveReinforcementLearning2022}).
}

We define the multi-objective return of a policy $\pi$ as a random variable
$\vect{G}^\pi \coloneqq \sum_{t=0}^\infty \gamma^t  \vect{R}_t$ from trajectories to vectors in $\mathbb{R}^m$, where the trajectories are drawn as $S_0 \sim \mu$, $A_t \sim \pi\left(\cdot \mid S_t \right)$, $\vect{R}_t = \vect{r}\left({S_t, A_t}\right)$, and $S_{t + 1} \sim p\left( \cdot \mid S_t, A_t\right)$ for all $t \geq 0$.
The multi-objective value function of a policy $\pi$ is the expected return, $\vect{v}^\pi(s) = \Ex_\pi[\vect{G}^\pi \mid S_0 = s]$.

\subsection{Rewards in MORL}\label{sec:rew-in-morl}

Knowing the multi-objective value functions of several policies
is not enough, by itself, to decide which policy should be preferred. The value of a policy is a vector: one component may improve while another one gets worse. To turn such trade-offs into a decision, MORL therefore requires two ingredients: a \emph{utility function} $u: \mathbb{R}^m \mapsto \mathbb{R}$, which assigns a scalar utility to a reward or return vector, and an \emph{optimisation criterion}, which specifies when this utility is applied.

The placement of $u$ matters. It determines whether we care about the utility of an average outcome, the average utility of whole trajectories, or the cumulative utility of each individual reward. These choices coincide for linear utility functions, but they can lead to different optimal policies when $u$ is non-linear. Two optimisation settings have been widely used in the MORL literature: Scalarised Expected Returns, and Expected Scalarised Returns.

\subsubsection{Scalarised Expected Returns}

In the scalarised expected returns (SER) setting, the goal is to find
\begin{equation}
\pi^* \in \arg\max_\pi u(\vect{v}^\pi) = \arg\max_\pi u(\Ex_\pi[\vect{G}^\pi]),
\end{equation}
given \textit{any} utility function $u: \mathbb{R}^m \mapsto \mathbb{R}$.
SER first averages the return vector of a policy and only then applies the
utility function. Intuitively, it evaluates a policy by its long-run mean
performance across objectives. This is natural when the same policy is used many times and the decision maker mainly cares about aggregate performance over a population of runs.

\subsubsection{Expected Scalarised Returns}

In the expected scalarised returns (ESR) setting, the goal is to find
\begin{equation}
\pi^* \in \arg\max_\pi \Ex_{\pi}\left[ u\left( \vect{G}^\pi\right) \right],
\end{equation}
given \textit{any} utility function $u: \mathbb{R}^m \mapsto \mathbb{R}$.
ESR applies the utility function to the complete return vector of each trajectory and only then averages across trajectories. It therefore captures how good an individual realised outcome is before taking expectations. This distinction is important when the utility function represents risk, fairness, saturation, or any other non-linear preference over complete outcomes.
%\lucas{Random thought: Does ESR make sense in infinite horizon problems as well? If not, perhaps mention it.} \dmr{Definitely, as long as the MOMDP is episodic. The only thing that truly makes sense otherwise would be average reward though - so not even SER either. }
%\lucas{\checkmark}

\subsubsection{Returns of Scalarised Rewards}
\label{sec:returns-scalar-rewards}
A third possible setting is the approach used by default in single-objective RL: apply the utility function directly to the reward vector at each time step, and then maximise the expected return of the resulting scalar rewards. In returns of scalarised rewards (RSR), the scalar return is
\begin{equation}
G_u^\pi = \sum_{t=0}^\infty \gamma^t  u(\vect{R}_t),
\end{equation}
and the goal is to find
\begin{equation}
\pi^* \in \arg\max_\pi \Ex_{\pi}\left[ G_u^\pi \right].
\end{equation}
RSR treats each time step as a separate trade-off. It is appropriate when the utility of an immediate reward vector is meaningful on its own, for example when one wants to penalise dangerous or undesirable instantaneous combinations of objectives, rather than only their cumulative effect over a full trajectory.

\subsubsection{Utility Functions}
If $u$ is linear (i.e., a simple weighted sum of the components of the reward or return vector), then the same policy will be optimal for SER, ESR, and RSR. However if a non-linear function is used for $u$ (e.g. Chebyshev distance), then a different policy may be optimal for each of these settings.
Importantly, this implies that we may have different optimal solution sets for each problem formulation \cite{hayes2022practical}.

\subsection{Rewards in Succesor Features-based RL}

The Successor Features (SFs) framework~\cite{Barreto+2017,barretoFastReinforcementLearning2020} was originally proposed to model multi-task RL problems, where each task is defined by a different reward function.
It assumes that the reward function of an MDP is given by a linear function over a set of $d$ \textit{features} $\vect{\phi}(s,a) \in \mathbb{R}^d$ : 
\begin{equation}
    r_{\w}(s,a) = \vect{\phi}(s,a) \cdot \w. \label{eq:SF}
\end{equation} 
The SFs of a policy $\pi$ represents the expected sum of the features: $\vect{\psi}^\pi(s,a) = \Ex_\pi [\sum_{t=0}^\infty \gamma^t \vect{\phi}_t | S_0 = s, A_0 = a]$.

Intuitively, we can see this setting as an instantiation of the RSR setting where the vector reward $\vect{R}_t$ is replaced by the feature vector $\phi_t$, and $u$ is a linear function $u_{\w}(\phi_t) = \phi_t \cdot \w$.
\cite{alegreOptimisticLinearSupport2022} showed that, in this case, SFs are equivalent to a multi-objective value function, and the SFs framework is equivalent to MORL under linear utility functions.

\subsubsection{Beyond Linear Rewards in SFs}

The expressiveness of SF-based approaches depends heavily on the definition of the features $\phi$. Unlike the vector rewards in MORL that are typically specified by domain experts, the features $\phi$ can instead be learned by an agent. 
For example, these features can be learned via optimization of supervised criteria, e.g., to approximate a set of reward functions \cite{Barreto+2018}, or via unsupervised objectives, e.g., to approximate the Laplacian of the MDP \cite{chandrasekarOptionBasisOptimize2025} or to identify clusters of the state space~\cite{bagot2025successor}.

Importantly, if the feature vector is a one-hot encoding indicating the current state of the agent in the state space, $\phi\left(s\right) = \left[\mathbbm{1}_{S_1}(s) ,\cdots, \mathbbm{1}_{S_d}(s)\right]^\top$ where $d = |\mathcal{S}|$, then any reward function, $r$, can be represented as a vector $\vect{w} \in \mathbb{R}^{|\mathcal{S}|}$, where $\vect{w} = [r(s_1) \ ... \ r(s_{|\mathcal{S}|})]^\top$.
This implies that non-linear functions of the reward vector (Section~\ref{sec:returns-scalar-rewards}) would be unnecessary.
The Forward Backward (FB) representation \cite{touatiLearningOneRepresentation2021,touatiDoesZeroShotReinforcement2022} fits in this context by learning a high-dimensional representation that, under some dimensionality constraints/assumptions, can express the optimal policy for any possible reward function. 

We observe a tendency for SF-based approaches to increase their expressiveness (the family of problems they can solve) by discovering more powerful feature representations, while maintaining a linear utility function. 

In contrast, in the MORL literature, most works typically consider a low-dimensional reward vector, but expand the family of problems that MORL techniques can solve by employing non-linear utility functions.
We believe that, although orthogonal, it makes sense to combine both directions when tackling some real-world problems.

That being said, the multi-objective RSR criterion with non-linear $u$,  can be implemented in successor features by simply appending the existing features $\phi_t$ by a non-linear combination of these features: $\phi_t \leftarrow \phi_t \cup \{\tilde{u}(\phi_t)\}$.\footnote{This assumes all features in $\phi_t$ are reward components. If this is not the case we would instead have to apply a composite function $u \circ \tilde{r}$, where $\tilde{r}$ is the translation of the successor features to the rewards in each objective.} This feature then gets accrued over time as the utility, and a weight of $1$ is placed on this new feature. SER and ESR in general however, cannot be implemented using traditional successor feature approaches, as after accruing rewards over time, only linear combinations are allowed.

%\flo{And what if we don't have a linear utility function? How should we interpret SFs in a multi-objective point of view, intuitively? I think this should be explained more in-depth here.}
%\lucas{I discussed this in a bit in the related work section. I expanded the section now.}

\section{Motivating Example: Toxicity}
\label{sec:toxic}
As mentioned above, the optimisation criteria (SER, ESR, and RSR) say something about the impact of---reward or feature---signals on user utility. This happens at different timescales: across multiple episodes for SER, over one episode for ESR, and at a single-timestep for successor features/RSR. All of these make sense in one setting or another. However, we would like to argue here, that they can also make sense all at once. It is just a matter of timescale.

Let us turn to the example of exposure to substances and their toxicity.  Toxicity effects typically follow a non-linear dose-response curve, and thus leads to non-linear utility. A high dosage in a short time can lead to severe immediate adverse effects. This is known as \emph{acute exposure} (one dose, or accumulating exposure up to 24 hours), and could correspond to a timestep-based application of a non-linear (utility function) before the rewards/penalties are added up at all.
%\flo{should we call this a reward? let's be clear, this is a penalty} %\peter{Maybe we say "before the rewards/penalties are added up"?}

However, just because a substance does not hurt you when you are exposed to it at a lower dosage once, does not mean you can keep getting exposed to it. In fact, in toxicology the following exposure time-lengths are often distinguished \emph{sub-acute} (24 hours to 28 days), \emph{sub-chronic} (less than 90 days), and \emph{chronic} exposure \cite{DENNY2024149}. 

Let's consider a small company with three certified people to work with toxic substances. These workers have different experience levels, and therefore also different risk exposure risks for different tasks.
%\flo{I think we must be careful because several conferences (e.g., that was the case for AAMAS) requires to include an ethical statement at the end of each paper if applicable. I'd say this is something we need to consider here.}
Let's imagine we aim to optimise the monthly work schedule, with an employee taking the same work station for a day, doing one task. For simplicity lets imagine there are 3 tasks in total that are the same each day. Let's take the dose each of the three employees get as three separate reward functions. What happens when we take different types of exposure into account?  

\subsection{Intuition}

First let us see what would happen if we would take different types of toxic exposure into account in an intuitive manner, before looking at a numerical example.

\paragraph{Chronic exposure.} For chronic exposure, we would consider the long-term damage of being exposed to a toxin in small amounts over a long period of time. This would mean we ought to look at the exposure over the course of multiple months or even years. In RL terms this would correspond to multi-objective RL (MORL) under the scalarised expected returns (SER) criterion \cite{hayes2022practical}. Furthermore, because the response to toxins are typically sigmoid shaped---meaning that the toxic effects start increasing slowly for low dosages and then increase ever faster before flattening off again---this means we probably just have to spread out the average toxic exposure over all employees as much as possible so that each individual's long-term average dose is low, leading to the least averse effects.

\paragraph{Sub-acute exposure.} For sub-acute exposure, we would be interested in the total dosage an employee would get over the entire month. This would require us look at the possible outcomes for each month, as a probability distribution, and look at the expected adverse effects of the cumulative dosages for the entire month (again by applying a probably sigmoidal response curve). It would probably also be important to track closely the dosage the employee already received up until a given day, and dynamically adapt our schedule to that. In RL terms this would correspond to expected scalarised returns (ESR) \cite{roijers2018multi}. 

\paragraph{Acute exposure.} For acute exposure, we would be interested in the immediate effects of a high dose within 24 hours. This would entail considering the dosage an employee gets as a result of a task, and then applying the response curve directly to that. Taking the average of that would correspond to the average risk of acute exposure on a daily basis. We could model this using successor features, by directly considering the toxic response---per timestep---as a state feature, corresponding to RSR. 

So what would a responsible employer have to do now? In the decision theoretic literature we seem to have the choice between SER, ESR, or RSR as an optimisation criterion. Each of these would imply only one of the different timescales for toxic exposure. But that does not suffice; a responsible employer should have their eye on \emph{every} type of toxic exposure, not just one. We should optimise for all at once. 

We argue that this indicates a critical gap in the literature, i.e., different problems may require us to look at (cumulative) rewards on different timescales, but importantly \emph{different timescales may have to be taken into account at the same time}. 

\subsection{Numerical example} \label{sec:example}
%\dmr{TODO: Liam with some help from Diederik}

Now we turn our example into a numerical decision problem, and will show that different policies are optimal for different settings.
%\begin{itemize}
%    \item Propose four different policies, three being the best out of the four for each individual sub-utility function, and one being the best for the complex combination
%\end{itemize}
The decision problem consists of 3 employees of varying skill levels and 3 tasks with varying toxicity (table \ref{tab:MDP}). The goal is to assign the employees optimally over time given a possibly non-linear utility function that limits the maximal toxic dosage per employee either per day, per month, over multiple months or over all possible time frames simultaneously. This problem can be modeled as a MOMDP with state space $\Sspace$ defined as all permutations of employees, with their positions indicating their assigned task (e.g. [emp1, emp2, emp3] indicates that employee 1 is assigned to task 1, employee 2 to task 2 and employee 3 to task 3) and action space $\Aspace$ as the finite discrete set of permutation indices such that $|\Aspace| = 3!$. Note that the state needs to be augmented with the accrued reward in order to accommodate non-linear utility under ESR and SER \cite{vamplew2024utilitybasedreinforcementlearningunifying,reymond_actor-critic_2023,roijers2018multi}. The vectorial reward function is defined as $\textbf{r}(s_t,a_t) = [-r_1(s_t,a_t), -r_2(s_t,a_t), -r_3(s_t,a_t)]$, where $r_i$ indicates the toxic dose that each employee is exposed to on a given day. These doses are defined as $r_i = \max(0, (1 - S_i) \cdot T_j + \varepsilon)$, with $T_j$ being the toxicity of the task currently assigned to employee $i$ and $S_i$ the corresponding skill level of that employee. The doses are slightly stochastic, subject to the noise $\varepsilon \sim \mathcal{N}(0,10)$. We note that in a successor feature formulation, the received dosages would be appended to the state, rather than given as a reward vector. We consider episodes of 30 days long to simulate work schedules over a month. 
\begin{table}[]
\caption{The environment parameters. Each employee initially starts at their respective task ($s_0 = [0,1,2])$.\label{tab:MDP}}
\begin{tabular}{ll}
\hline
           & Skill level \\ \hline
Employee 1 & 0.75        \\
Employee 2 & 0.60        \\
Employee 3 & 0.2         \\ \hline
\end{tabular}
\quad
\begin{tabular}{ll}
\hline
       & Toxicity \\ \hline
Task 1 & 70       \\
Task 2 & 25       \\
Task 3 & 10       \\ \hline
\end{tabular}
\end{table}

In order to model and mitigate the three types of toxic exposure highlighted above, we propose three utility functions. \paragraph{The acute utility function} determines the limit on acute exposure and as such, should be applied to the reward at each timestep:
\begin{equation}
u_{acute}(\textbf{r)} = \sum_{i=1}^3\frac{1}{1+e^{-.5(\textbf{r}_i + 25)}} \label{eq:dailyU}    
\end{equation}
\paragraph{The episodic utility function} considers sub-acute and chronic exposure (e.g. over one or multiple episodes) and thus should be applied over the returns or values:
\begin{equation}
u_{episodic}(\textbf{G}_t) = \sum_{i=1}^3\frac{1}{1+e^{-.02(\textbf{G}_{t,i} + 400)}} \label{eq:epU}
\end{equation}
Note that in the case of SER optimization $u_{episodic}(\textbf{V}^\pi)$ should be computed instead of over the return $\textbf{G}_t$ as under ESR.

\paragraph{The combined utility function} should be a combination of both $u_{acute}$ and $u_{episodic}$ (applied to both the returns and the value). This is a novel approach to solving multi-objective sequential decision-making problems using a utility-based approach.

\section{Experiments}
To showcase the need for a unified view we performed experiments on the example highlighted in section \ref{sec:example} for the three optimisation criteria: ESR, SER and RSR. Finally, we combined the three criteria through a multi-objective natural evolution strategy (MONES) \cite{glasmachers2010mones,salimans2017evolutionstrategiesscalablealternative}. Note that the experiments serve as an illustrative proof of concept to show distinct differences between optimisation criteria.
\subsection{Algorithms}
In order to show the differences in policies learned under each optimization criterion, we used algorithms that are tailored for those settings \footnote{The code for our experiments is available at \url{https://github.com/liammertens/MORL-timescale/tree/main}}.
\paragraph{EUPG (ESR)} Expected Utility Policy Gradient (EUPG) \cite{roijers2018multi} is a multi-objective adaptation of the REINFORCE algorithm \cite{williams_simple_1992}. It is designed to learn optimal policies under ESR for non-linear utilities and achieves this by incorporating the accrued reward during learning and applying the utility over the sum of the accrued return and future returns.
\paragraph{NLPPO (SER)} Optimising for non-linear utility under the SER criterion requires computing the value function first before computing the utility function. This is achieved by modifying an offline variant of proximal policy optimisation (PPO) \cite{schulman2017proximalpolicyoptimizationalgorithms,meng_off-policy_2023} to apply the utility over the critic network's output, as proposed by \cite{ropkeDivideConquerProvably2025b}.

\paragraph{SFDQN (SF/RSR)} We simulated a successor feature (/RSR) approach for this problem by transforming the problem into a linear multi-objective RL problem through the application of $u_{acute}$ (Eq. \ref{eq:dailyU}), which is a non-linear transformation with a subsequent linear combination using uniform weights (all employees counting equally) (as in Eq. \ref{eq:SF}). This non-linear transformation of the reward function corresponds to the learning of successor features that are non-linear w.r.t. the states in the problem, given that the original state space were extended with a non-linear utility applied to the reward components. After application of the linear scalarisation on the non-linear features, we are left with a single-objective RL problem that can be solved using deep Q networks (DQN) \cite{mnih2013playingatarideepreinforcement}.

\paragraph{MONES (combined perspective)} Optimising for the combined utility over different timesteps requires a novel approach, as directly combining either of the methods above is infeasible because scalarisation happens over different time windows and does not distribute over time due to the non-linearity of the utilities. We propose to use a natural evolution strategy to learn the search distribution over policy parameters that optimises the combined utility per timestep, per episode and on average over multiple episodes. This combination is achieved by multiplying all three utilities to get the fitness function of the parameter population, inspired by the AND operator in fuzzy logic. In order to ensure equal contribution of each utility to the total, we divide $u_{acute}$ by 30, the episode length, to get the average acute utility. This leads to equal contribution of each sub-utility to the total due to the normalisation effect of the sigmoidal utility functions.

\subsection{Setup}
The hyperparameters for the algorithms are given in table \ref{tab:hyperparameters}. Each experiment was performed using 3 different seeds on an Apple M5 CPU.
\begin{table}[ht!]
    \centering
    \caption{Hyperparameter configurations for the evaluated algorithms.}
    \label{tab:hyperparameters}
    \small
    \begin{tabular}{lr} 
        \toprule
        \textbf{Hyperparameter} & \textbf{Value} \\
        \midrule
        \multicolumn{2}{c}{\textbf{Shared}} \\
        \midrule
        Discount factor ($\gamma$) & 1 \\
        Episode length & 30 \\
        Optimizer & Adam \\
        \midrule
        \multicolumn{2}{c}{\textbf{EUPG (ESR)}} \\
        \midrule
        Learning rate & 1e-4 \\
        Replay buffer size & 1e5 \\
        Learning steps & 3e6 \\
        \midrule
        \multicolumn{2}{c}{\textbf{NLPPO (SER)}} \\
        \midrule
        Actor learning rate & 2.5e-4 \\
        Critic learning rate & 2.5e-4 \\
        Clip range ($\epsilon$) & 0.2 \\
        Entropy coefficient & 0.05 \\
        GAE $\lambda$ & 0.95 \\
        Learning steps & 3e6 \\
        \midrule
        \multicolumn{2}{c}{\textbf{SFDQN (RSR)}} \\
        \midrule
        Learning rate & 1e-4 \\
        Replay buffer size & 1e5 \\
        Target net update freq. & 1000 \\
        Exploration fraction & 0.5 \\
        Learning steps & 2e6 \\
        \midrule
        \multicolumn{2}{c}{\textbf{MONES (Combined)}} \\
        \midrule
        Population size & 50 \\
        Eval episodes per member & 60 \\
        Learning rate & 1e-4 \\
        Noise std. dev. ($\sigma$) & 0.1 \\
        Learning steps (generations) & 5e6 \\
        \bottomrule
    \end{tabular}
\end{table}

\subsection{Results}
\begin{figure*}
    \centering
    \includegraphics[width=\linewidth]{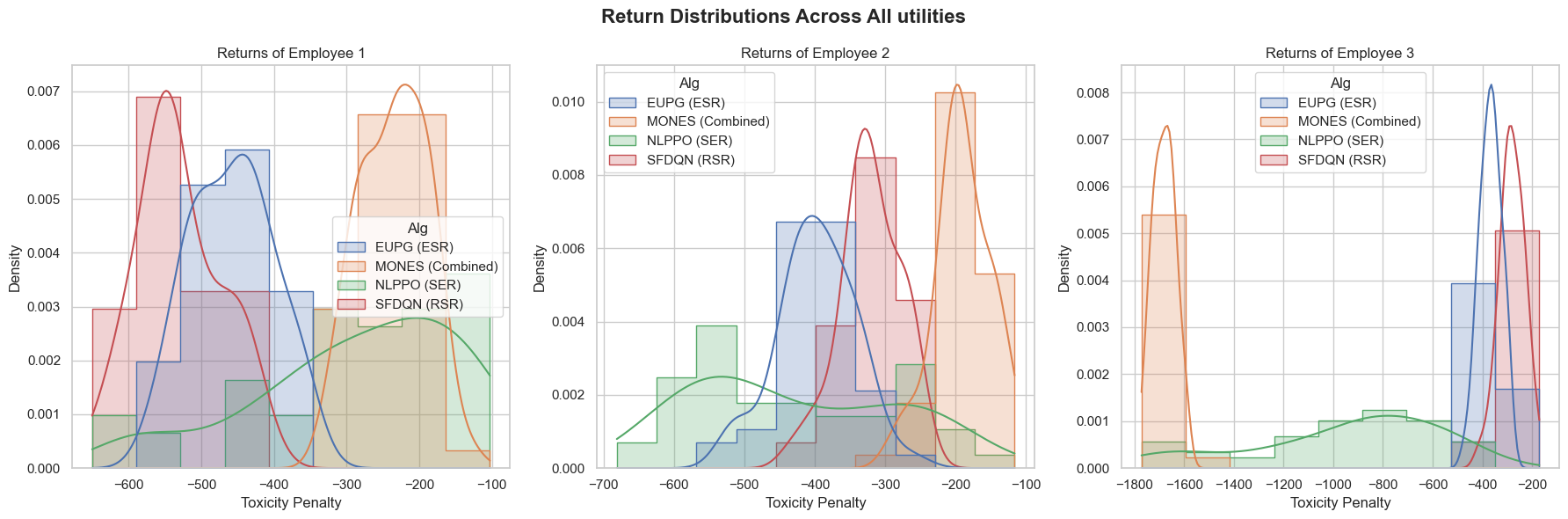}
    \caption{Comparison of the return distributions resulting over 50 policy rollouts.}
    \label{fig:results}
\end{figure*}
When comparing the return distributions of each method in Fig. \ref{fig:results}, we could notice differences between assignments. Optimising under ESR led to a switching strategy being learned that kept the toxicity penalty nicely around the threshold of 400 for all employees. Note that this implies exceeding the daily intake limit at times. Looking at the optimal policy under SER, the return distribution has less kurtosis. This is the result of the policy preferring to keep employees 1 and 2 under the threshold by rotating them, but in turn 'sacrificing' the least experienced employee 3.\footnote{Please note that this is a numerical example problem, and we would not recommend this in practice.} By maximising the utility for 2 employees instead of nicely balancing, it aims to get a higher total utility. Note however, that the expected return under SER lies close to the expected return under ESR for employee 1 and 2, although with increased variance. This shows why SER optimisation is not optimal in cases where risk-awareness is necessary. The need for sacrificing employee 3 stems from the rollouts where employee 1 and 2 are kept well under the toxicity threshold for the episode, leading to optimal \textit{expected} utility. Optimisation under RSR leads to a switch-averse policy that prefers to keep employees working on the tasks that correspond to their skill level, as this strategy is certain to keep working and no additional requirements on total toxicity are given. Finally, the combined method learns to shuffle, but mainly prefers to put employee 3 on the most difficult task, even more than under SER, leading to optimisation for the utilities of the two most experienced employees. One explanation for this phenomenon could be that the combination of all 3 utilities allows the agent to maximise one or two of them and neglect the other, leading to the 'sacrifice' scenario. E.g. the agent is not willing to risk exceeding the daily thresholds for employee 1 and 2 in exchange for the total utility of employee 3, so they are kept on the less intensive tasks.

The sacrificing policies being optimal for SER and the combined objective are a clear indication that the problem is too hard, as not all employees can be safe from chronic exposure. Especially in the combined objective, it appears that the safety of the third employee is given up, indicating a very clear safety issue for the employer. However, if you look at the RSR or ESR optimisation, the switching policies do seem to balance the toxicity well, and the employees are safe. As such, there seems to be a lot of value in looking at the different timescales at once, if we want to guarantee the safety of all. Clearly the employer needs more employees, or better hazard mitigation measure, to lower the exposure risks when looking at the combined objective, while they might not think so looking solely at ESR (or RSR). In this---admittedly highly simplified/abstract---example, looking at different timescales at once is thus truly \emph{essential}.

To conclude we observe that there are problems for which looking at different timescales individually leads to very different results than looking at the combined utility effects on different timescales simultaneously. For this we used three standard algorithms for RSR, ESR and SER, and a heuristic approach (MONES) for the combined utility of different timescales. We believe that these results call for further investigation into, and specialised algorithms for, combined timescale utility optimisation.

\section{Related Work}

The majority of MORL algorithms consider only a single optimisation criteria. For example, \cite{caiDistributionalParetooptimalMultiobjective2023} shows how to, given an MOMDP under ESR, construct a MDP with augmented state space and  scalar reward function that induces the same optimal policy as the original MOMDP. %However, this requires conditioning the state of the agent with accumulated returns and time step.
\cite{vamplew2022impact} discussed a scenario where the optimal policy might maximise SER subject to a satisfactorily small level of variation in returns between episodes (reduced-variance SER), but their proposed algorithm is restricted to the specific case of finding SER-optimal deterministic policies for environments with stochastic rewards but deterministic state dynamics. More recently \cite{Ropke+2023} introduced the distributional undominated set (DUS), which in theory contains all optimal policies under both SER and ESR criteria.

At a theoretical level, studies have examined how the choice of utility function or optimisation criterion impacts on the policies that are available to a RL or MORL agent. \cite{subramaniExpressivityObjectiveSpecificationFormalisms2024} proposes a hierarchy of RL formalisms, and shows that, in theory, the SER (Outer MORL) formalism can express all policy orderings that ESR (Inner MORL) can express. \cite{rodriguez-sotoAnalyticalStudyUtility2024} shows a characterisation of the utility functions for which an associated optimal policy exists, and (2) a characterisation of the types of preferences that can be expressed as utility functions.

%\peter{I've done some restructuring of this Related Work section to make it more a discussion of topics rather than just a list of papers. But I'm not as familiar with the other papers below. Lucas, I i think you added these so can you try to tie this together?}
%\lucas{Yes!}

In the literature on multi-task RL and successor measures/features, a few works have explored similar problems.
Regarding approaches that learn distributional value functions,
\cite{wiltzerDistributionalAnalogueSuccessor2024a} introduces a distributional analogue to the SFs. In particular, they show that it allows agents to perform zero-shot \textit{risk-sensitive} policy evaluation, which can be seen as a type of non-linear utility.
The Forward-Backward (FB) representation~\cite{touatiLearningOneRepresentation2021,touatiDoesZeroShotReinforcement2022} is another relevant framework that allows to approximate (under some dimensionality constraints) the optimal policy for any scalar reward function, fitting the formalism presented in Section~\ref{sec:returns-scalar-rewards}.
Recently, \cite{bagatellaSoftForwardBackwardRepresentations2026} studies how to optimize for general non-linear utility functions of the successor measure of a policy. We believe this type of method is a good example of combining expressive reward representations with non-linear utilities.

%\dmr{TODO: Option Keyboard, Lucas?} \lucas{\checkmark}
Another relevant line of work that extends the expressivity of SF-based methods is the Option Keyboard (OK)~\cite{barretoOptionKeyboardCombining2019a,barretoFastReinforcementLearning2020}. Instead of following a policy specialized to a fixed preference vector $\w$, an OK agent is capable of dynamically changing this vector (and the associated policy being followed) at each time step, similar to dynamic preferences in MORL \cite{abels_dynamic_2019}.
This type of technique has been shown to be capable of optimally solving some classes of non-linear reward functions~\cite{alegreConstructingOptimalBehavior2025}.

\section{Discussion}
\label{sec:disc}

In this paper, we have shown the need for a utility-based perspective in multi-objective planning and reinforcement learning that works on multiple timescales at once. We have shown this intuitively and then numerically on a toxicity example. 

In our example---for reasons of simplicity---we left out the sub-chronic toxicity that is also known from the toxicology literature. In terms of timescale, this would fall between SER and ESR; it goes beyond one execution of the policy (encompassing about a month), but would not be quite the average longtime outcome either. There is some discussion over the exact timescale of what it means to be sub-chronic\footnote{See e.g., %\url{https://www.plantchemsci.com/posts/the-hidden-language-of-toxicity-tests-why-a-28day-study-isnt-subchronic}
\url{https://tinyurl.com/subchronic}.}, but if we take the 90 days mark it would correspond to roughly three policy executions in our example. We could look at the distribution of the summed doses in three policy executions, which would be a smoother distribution of values. This could be added as a fourth timescale.

Although we only used one abstract example in this paper, we do believe that these types of problems are actually quite common. Consider for example companies that are interested in the average time a customer's order needs to be completed, in order to optimise serving costs. However, they might also be interested in the individual customer's waiting times at the same time, as if a handful of customers might have to wait very long compared to the main bulk, this can lead to very bad reviews and lead to damage to the company's reputation. This would suggest at least a combined SER-ESR perspective and corresponds to an average reward RL problem \cite{mahadevan_average_1996}. In the medical domain (possibly AI-assisted) radiation therapy \cite{van2019benefits} is spread out over multiple sessions to reduce the radiation damage to surrounding organs, while making sure the accumulated dose over time is enough to kill a tumor. This suggests a combined RSR-ESR perspective. So indeed, we suspect that a lot of critical real-world decision problems exhibit utility properties on multiple timescales.  

With the numerical example, we have shown that different policies work better for different timescales and the corresponding optimisation criteria. We argued that we need to develop comprehensive methodology and algorithmic approaches to deal with different timescales simultaneously, as policies that work well for SER, ESR, or RSR may not suffice for a combination of the three. 

We believe that a key aspect of any solution that works on multiple timescales simultaneously will be a distributional one, i.e., to learn a distributional return and distributional reward function. However, we do note that even if we have those, policy optimisation is far from trivial. A change in policy can have effects on multiple timescales, leading to a highly complex utility landscape. The AI community has developed multiple ways to tackle such complex utility landscapes though, including policy gradients, evolutionary strategies, etc. So we are hopeful that our community can and will find good approaches for this issue.

\appendix

\section*{Ethical Statement}
%\dmr{TODO: Peter?}
%\peter{\checkmark}
The toxic exposure example in this paper is used purely for illustrative purposes. We are not advocating for organisations to manage employees' exposure to toxins in this manner.

More broadly, we believe that the greater expressivity provided by the use of multiple rewards and optimisation criteria enhances the AI agents' capacity to include ethical considerations explicitly during decision-making \cite{vamplew2018human}. 

\section*{Acknowledgments}
We acknowledge the following funding: LPHM was supported by the VUB; FD was supported by the Belgian Flemish AI Research Program and the “DESCARTES” iBOF project; DMR was supported by the PEER project, which has received funding from the EU’s Horizon Europe Research and Innovation Programme, under Grant Agreement number $\#101120406$.  
% Lucas:
This study was financed in part by the Coordenação de Aperfeiçoamento de Pessoal de Nível Superior - Brasil (CAPES) - Finance Code 001. This work was supported by Kunumi Institute. The authors thank the institution for its financial support and commitment to advancing scientific research.

We would like to thank Siri Willems (of imec AI labs) for her input and feedback on the medical examples.

\bibliographystyle{named}
\bibliography{ijcai26,lucas}

\end{document}